\documentclass{article}

\usepackage{arxiv}
\usepackage{tikz}
\usepackage{bbold}
\usetikzlibrary{calc}
\usepackage{tikz-cd}
\usepackage[utf8]{inputenc} 
\usepackage[T1]{fontenc}    
\usepackage{hyperref}       
\usepackage{url}            
\usepackage{booktabs}       
\usepackage{amsfonts} 
\usepackage{amsmath}
\usepackage{amssymb}
\usepackage{nicefrac}       
\usepackage{microtype}      
\usepackage{lipsum}		
\usepackage{graphicx}
\usepackage{natbib}
\usepackage{doi}
\usepackage{caption} 
\title{Causal Generative Explainers using Counterfactual Inference: A Case Study on the Morpho-MNIST Dataset}

\author{\normalfont Will Taylor-Melanson\thanks{Email: wl647481@dal.ca}\\
	Faculty of Computer Science\\
	Dalhousie University\\
	Halifax, Nova Scotia, B3H 1W5, Canada\\
	\and
        Zahra Sadeghi\\
	Faculty of Computer Science\\
	Dalhousie University\\
	Halifax, Nova Scotia, B3H 1W5, Canada \\
        \and
        Stan Matwin\\
	Faculty of Computer Science\\
	Dalhousie University\\
	Halifax, Nova Scotia, B3H 1W5, Canada
}

\date{}

\renewcommand{\shorttitle}{Causal Generative Explainers using Counterfactual Inference}

\hypersetup{
pdftitle={Causal Generative Explainers using Counterfactual Inference: A Case study on the Morpho-MNIST Dataset}
}

\begin{document}
\maketitle

\begin{abstract}
In this paper, we propose leveraging causal generative learning as an interpretable tool for explaining image classifiers. Specifically, we present a generative counterfactual inference approach to study the influence of visual features (i.e., pixels) as well as causal factors through generative learning. To this end, we first uncover the most influential pixels on a classifier's decision by varying the value of a causal attribute via counterfactual inference and computing both Shapely and contrastive explanations for counterfactual images with these different attribute values. We then establish a Monte-Carlo mechanism using the generator of a causal generative model in order to adapt Shapley explainers to produce feature importances for the human-interpretable attributes of a causal dataset in the case where a classifier has been trained exclusively on the images of the dataset. Finally, we present optimization methods for creating counterfactual explanations of classifiers by means of counterfactual inference, proposing straightforward approaches for both differentiable and arbitrary classifiers. We exploit the Morpho-MNIST causal dataset as a case study for exploring our proposed methods for generating counterfacutl explantions. We employ visual explanation methods from OmnixAI open source toolkit to compare them with our proposed methods. By employing quantitative metrics to measure the interpretability of counterfactual explanations, we find that our proposed methods of counterfactual explanation offer more interpretable explanations compared to those generated from OmnixAI. This finding suggests that our methods are well-suited for generating highly interpretable counterfactual explanations on causal datasets.

\end{abstract}


\section{Introduction}
Deep Generative Models (DGMs) are statistical models which aim to generate new synthetic samples following the distribution of training sets using deep neural networks. The recent advancement in DGMs' architecture and learning techniques have resulted in a new trend in Artificial Intelligence (AI) known as generative AI with a wide variety of applications in different industry sectors such as art, healthcare and marketing. Two popular types of DGMs are Variational Autoencoders (VAEs)~\citep{rezende2015variational} and Generative Adversarial Networks (GANs)~\citep{goodfellow2020generative}.

Recently, authors have proposed pairing DGMs with causal mechanisms to form structural causal models (SCMs) over high-dimensional datasets~\citep{pawlowski2020deep,dash2022evaluating}. We refer to such models as Causal Deep Generative Models (CGMs) for convenience, as they are deep generative models which preserve assumed or revealed causal constraints between variables of interest. We note that this acronym conflicts with the causal graphical models discussed by~\cite{scholkopf-statistical-to-causal}, though causal graphical models are not otherwise discussed in this work. While CGMs have been applied to data generation and detection of bias in image classifiers~\citep{dash2022evaluating}, methods using CGMs to explain image classifiers have thus far been limited to binary classifiers and binary attributes. In this paper, we investigate expanding the applicability of CGMs to Explainable AI (XAI). XAI refers to methods that offer human understandable interpretations for complex machine learning models. XAI techniques help to enrich the capacity of AI systems by providing understanding, transparency, fairness, and trust for the decisions made by AI systems~\citep{dwivedi2023explainable}.

This paper explores explainability methods for image classifiers by focusing on the utilization of two causal deep generative architectures, namely, the DeepSCM~\citep{pawlowski2020deep} and ImageCFGen~\citep{dash2022evaluating} models. The authors of ImageCFGen defined a counterfactual importance score using their BiGAN-based architecture, which measures how a classifier's prediction changes when a given attribute (i.e., cause of the classifier's input) changes. For instance, measuring how an “attractive” classifier's output changes when the attribute “bald” changes. While this method proved useful in the work by Dash et al., it is not without limitations. Namely, that it is only defined for binary attributes and binary classifiers. These limitations motivate the methods described in this paper, which aim to provide image-based and attribute-based classifier explanations on datasets with continuous and categorical variables of interest representing the causes of an image. The novelty of these methods lies in the fact that they are easily implemented using any pretrained causal generative model such as DeepSCM and ImageCFGen, rather than requiring explanation-specific generative models trained with custom loss functions for individual classifiers~\citep{van2021conditionalcf, yang2021cfmodel}. The source code for our experiments is available in a public Github repository.\footnote{\url{https://github.com/wtaylor17/CGMExplainers}}

The first method we propose exploits generative learning for deriving model explanations, contrasting model behaviour as the value of an attribute varies. Specifically, we use SHAP values~\citep{shap} and contrastive explanations~\citep{dhurandhar2018contrastive} to visualize the importance of pixel features as a causal generative model varies the attributes of a causal dataset via counterfactual inference.

The second method proposed in this paper uses Monte-carlo sampling in the latent space of conditional generative models to assign classification scores to attribute values rather than image values, allowing feature importance methods such as SHAP for structured data to be applied to image classifiers in settings where causal metadata is available in image datasets.

The third and final method, counterfactual explanation, uses the BiGAN and VAE architectures of ImageCFGen and DeepSCM respectively to generate counterfactual explanations which are realistic and interpretable. We use quantitative metrics to compare our proposed methods to an open-source toolkit for model explanation, demonstrating the viability and interpretability of our counterfactual explainers.

\section{Related Work}

\subsection{Causal Generative Models}

Work from~\cite{pawlowski2020deep} was the first to use deep, non-invertible mechanisms to learn high-dimensional flows for counterfactual inference. Referred to as DeepSCM, the work proposes three ways to model the data generation process of an SCM while allowing for counterfactual inference. The mechanism most relevant to this work is the ``amortized, explicit" strategy, which is most commonly modelled with a VAE. The use of an encoder-decoder model is to allow for \textit{abduction}, the recovery of unobserved noise variables from observed data, a necessary part of computing a counterfactual~\citep{scholkopf-statistical-to-causal}. The low-level attributes of a causal dataset (i.e., non-image variables) are modelled for the DeepSCM algorithm via the method of normalizing flows~\citep{kobyzev2020normalizing}, which uses a change of variables to make the density of a transformed variable tractable. In their experiments, Pawlowski et al. perform counterfactual inference on the Morpho-MNIST dataset~\citep{morpho-mnist}, a modified version of the famous MNIST handwritten digits dataset\footnote{The original MNIST dataset may be downloaded from~\url{http://yann.lecun.com/exdb/mnist}.}~\citep{deng2012mnist} with added morphological attributes of thickness, intensity, and slant. Because the Morpho-MNIST dataset has its morphological operations readily available for programmers, any causal structure can be assumed over its attributes and implemented in a man-made causal graph, as was done by the authors of DeepSCM. Because of this it is straightforward to evaluate the accuracy of interventions performed on the Morpho-MNIST attributes, which showed that the DeepSCM model has high accuracy in manipulating the thickness, intensity, and slant of an image in a causally correct manner. While Pawlowski et al. have success in training accurate causal generative models using VAEs, they do not directly experiment with using these models to explain image classifiers.

The ImageCFGen algorithm from~\cite{dash2022evaluating} is another recently proposed causal generative model for counterfactual inference. However, rather than using a VAE, a bidirectional GAN (BiGAN)~\citep{bigan,ali} is used to model the data generation process. This has the advantage of a fully-deterministic model, i.e., latent vectors do not come from a distribution but are rather computed directly by an encoder both during training and inference time. However, recent theoretical results have also suggested that models such as deterministic BiGANs are not universal approximators in the usual sense, and may not be stable during training in practice~\citep{feng2021uncertainty}. Despite this, the ImageCFGen algorithm achieved comparable performance to the DeepSCM algorithm on the Morpho-MNIST dataset during experimentation in the work by Dash et al. To aid in the reconstruction of images and the approximation of counterfactuals, Dash et al. also proposed a fine-tuning procedure for their model which was shown to improve the preservation of image style during counterfactual approximation.

\subsection{Explaining Classifiers with Generative Models}
While Pawlowski et al. do not directly propose a method of classifier explanation using their causal generative model, Dash et al. propose a basic method of measuring the importance of a binary attribute to a binary classifier, which they call their \textit{counterfactual importance score}. They do this by measuring the difference in a classifiers output when performing a counterfactual changing the binary attribute from one value to its other value. For instance, when measuring the importance of the attribute \textit{bald} to an attractiveness classifier on a dataset of human faces, it was found by Dash et al. that the classifier associated the presence of baldness with a lower probability of attractiveness. Using a similar method, Dash et al. were able to measure the bias of a binary classifier with respect to a binary attribute. The main downside of these methods is that they only work in the very specific setting where both the attribute of interest and the output of the classifier being explained are binary.

Another approach to explaining classifiers using causal models is the explainer described by~\cite{parafita2019explaining}, who use a causal model to measure the effect of ``latent factors" (what we refer to as attributes or causes of an image) on a binary classifier. However, their approach does not explicitly perform abduction and thus does not produce counterfactual data in the standard causal sense~\citep{pearl2009causality} as DeepSCM and ImageCFGen does. In fact, they avoid explicitly training a generative model in their experiments altogether, and instead simply look for examples in their synthetic dataset matching the desired attributes. As such, we don't consider the method from Parafita and Vitria to be a CGM in the same sense as the deep models used in this work.

A method proposed by~\cite{hvilshoj2021ecinn} uses invertible neural networks (normalizing flows)~\citep{dinh2014nice} to generate counterfactual examples. The invertible network acts as a generative classifier which is able to map input images to a latent space. This method proved to generate realistic counterfactuals, and requires very little computation to generate them. However, the method depends on the classifier being explained being an invertible NN, i.e., it cannot be used on an existing classifier.

\cite{mahajan2019preservingcf} propose counterfactual explainers which preserve causal constraints on variables in a dataset through the use of a structural causal model (SCM). However, the results from this work are limited to structured datasets, and the generative model used in their method is not directly responsible for generating model explanations.

\cite{van2021conditionalcf} propose using conditional generative models for the creation of counterfactual explanations. While the model is efficient and only requires forward passes to compute the counterfactuals, the trained model is specific to a certain classifier, meaning that a new generative model needs to be trained for each classifier to be explained. A similar approach comes from~\cite{yang2021cfmodel}, who train a conditional generative model  using an existing classifier to produce counterfactual explanations, optionally accounting for causal relationships by modifying their generator structure, but who do not report experiments on image data or otherwise high-dimensional datasets. While either of these methods may produce believable counterfactuals in an efficient manner, neither of them perform abduction, i.e., they do not use a general-purpose CGM to produce counterfactual data and cannot explain arbitrary classifiers without retraining.

\section{Methods}
\label{sec:methods}
In this section, we propose a novel approach of \textit{causal generative explainers}. More specifically, we propose three methods for explaining AI classifiers using causal generative models performing counterfactual inference. In the first method, we use counterfactuals with different levels of a given attribute for visualizing the evolution of pixel importances using SHAP~\citep{shap}, as well as the evolution of pertinent positives (PP) and pertinent negatives (PN) using a contrastive explainer~\citep{dhurandhar2018contrastive}. In the second method, we define a function over attributes to provide feature importances in the attribute space for an image classifier using SHAP. In the third method, we use CGMs to provide both gradient-based and model-agnostic counterfactual explanations of classifiers.

Before we begin, we restate the method of counterfactual approximation using causal generative models as described by the authors of ImageCFGen~\citep{dash2022evaluating}. To generate a counterfactual using a causal generative model with encoder $E$ and generator/decoder $G$, we begin with an observational pair $\mathbf{x,a}$ of an image and its attributes. We then ask the question ``what would $\mathbf{x}$ have been, if $\mathbf{a}$ had instead been $\mathbf{a}'$?". This requires us to recover the unobserved variable $\boldsymbol{\epsilon} \approx E(\mathbf{x,a})$ that was used to generate the original data via the encoder, and then to use the counterfactual attributes $\mathbf{a}'$ to construct a new image. The unobserved variable accounts for the randomness in the data generation process. For Morpho-MNIST, we interpret this as meaning that $\boldsymbol{\epsilon}$ accounts for the writing style of a handwritten digit. In symbols, the counterfactual $\mathbf{x}'$ is of the form:
\begin{equation}
    \label{eq:cgm_cf}
    \mathbf{x}' = G(E(\mathbf{x,a}),\mathbf{a}')
\end{equation}
see the paper from~\cite{dash2022evaluating} for a mathematical proof of the causal validity of Equation~(\ref{eq:cgm_cf}). Intuitively, for Morpho-MNIST, Equation~(\ref{eq:cgm_cf}) says to generate an image $\mathbf{x}'$ in the same style as $\mathbf{x}$, but with the attributes $\mathbf{a}'$ instead of $\mathbf{a}$. When the original observation $\mathbf{x},\mathbf{a}$ is fixed and the counterfactual attributes $\mathbf{a}'$ vary, we may denote the counterfactual from Equation~(\ref{eq:cgm_cf}) as $\mathbf{x}'(\mathbf{a}')$, a function of the counterfactual attributes, rather than writing out the entire right hand side of Equation~(\ref{eq:cgm_cf}).

\subsection{Pixel Explanations}

\label{sec:shap_contrastive_methods}
In our approach, we consider the attributes thickness, intensity, and slant of the Morpho-MNIST~\citep{morpho-mnist} dataset and aim to measure the influence of these attributes on an image classifier $f$ in the pixel space. We denote $\text{SHAP}(\mathbf{x};f)$ as the SHAP values (feature importances) from~\cite{shap} of a function $f$ on an input $\mathbf{x}$. For an input $\mathbf{x}$, $\text{SHAP}(\mathbf{x};f)$ consists of $K$ feature importance masks (saliency maps) with the same shape as $\mathbf{x}$, where $K$ is the number of classes predicted by $f$.

In the pixel space, we consider the SHAP values for various levels of a given attribute, in order to determine how the regions highlighted as positively or negatively impacting the output score of a classifier vary along with the attribute in question. As with any counterfactual, we begin with an observation $\mathbf{x},\mathbf{a}$, where $\mathbf{x}$ is an image and $\mathbf{a}$ are its corresponding attributes. To explain the classification $f(\mathbf{x})$ for a classifier $f$, we can compute $\text{SHAP}(\mathbf{x};f)$ as described by Lundberg and Lee. More specifically, we use the gradient explainer available in the publicly available implementation of SHAP.\footnote{\url{https://github.com/shap/shap/blob/master/shap/explainers/_gradient.py}} However, in this section we are interested in the consistency of the pixel feature importances as a given attribute $\mathbf{a}_i$ (one of thickness, intensity, or slant for Morpho-MNIST) is varied via counterfactual approximation. Denoting  $\mathbf{a}^{i,v}$ as the counterfactual attributes under the intervention $do(\mathbf{a}_i=v)$, the counterfactual image for this intervention is denoted $\mathbf{x}'(\mathbf{a}^{i,v})$ according to our previously established notation.
To formalize our visualization technique, we have a set $S_i = \{v_1, v_2, \ldots, v_N\}$ of values for the attribute $\mathbf{a}_i$, and wish to visualize the consistency of pixel SHAP values as we vary $\mathbf{a}_i$ through this set of values. The explanations given by SHAP are denoted as $\text{SHAP}(\mathbf{x}'(\mathbf{a}^{i,v_j});f)$ for each $v_j\in S_i$. We plot these pixel importance values along a vertical axis in order to visualize how the pixel importances evolve as the value of an attribute varies (see for instance~\autoref{fig:thickness_image_shap_vae}). When we generate counterfactual Morpho-MNIST digits, using the same latent vector $E(\mathbf{x},\mathbf{a})$ for each counterfactual ensures that the style is preserved between images. Thus, the motivation for using causal generative models in this experiment is to isolate the effect of the attribute in a causally correct manner, i.e., not also varying the style of the image. 

This section presents another perturbation based method for displaying influential pixels for different levels of a given attribute. However, instead of only displaying pixel importances for the counterfactuals corresponding to each attribute level, we visualize pertinent negatives (PN) and pertinent positives (PP) using the contrastive explainer from~\cite{dhurandhar2018contrastive}. This contrastive explainer method (CEM) illustrates two minimal sets of features, pertinent positives (PP) and pertinent negatives (PN) that finds the necessary pixel features to be, respectively, either present to produce the same classification or changed to produce a different classification. This explainer provides human-interpretable results and highlights the positively important and negatively important regions of an image to explain a classifier. The optimization problems corresponding to PN and PP explanations are available in the original contrastive explainer paper. We also choose to visualize the distribution of class scores $f_j(\mathbf{x})$ of the classifier to analyze how they evolve as a given attribute changes. Unlike with the method using SHAP described previously, this method provides two visual explanations (a PN and a PP) for any image $\mathbf{x}$, instead of pixel importances for the individual class scores. This method of visualization is in part meant to detect what types of attribute changes may act as adversarial to a classifier by changing its decision, or otherwise impacting its class score distribution. Further, this method is used to measure how consistent the positive and negative regions of the contrastive explanations are across counterfactuals with one varied attribute, i.e., how specific the explanations are to the original image compared to the style of the image itself.

\subsection{Attribute Explanation}
\label{sec:attribute_explanations}
\begin{figure}[!htbp]
    \centering
    \includegraphics[width=\textwidth]{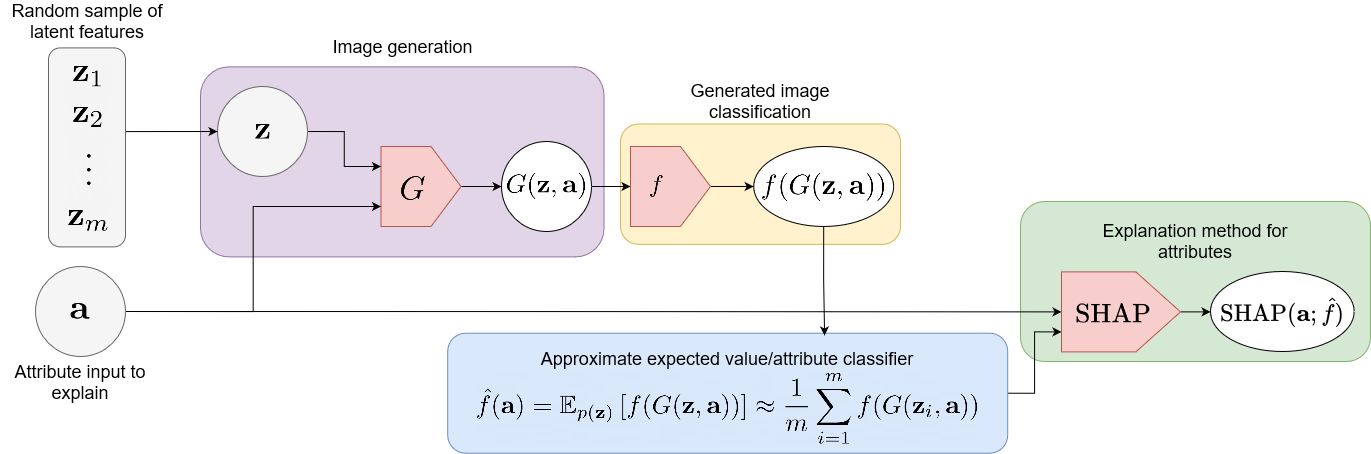}
    \caption{The proposed method of using a generative causal model (such as the generator $G$ of a BiGAN from ImageCFGen~\cite{dash2022evaluating}) to rank important human-interpretable attributes for an image classifier. Attributes to be explained are either sampled from an SCM $\mathcal{M}$ or from an available test set, and these along with a random sample of latent features form the input to the image generator $G$, which feeds inputs to a classifier $f$ to define a classification of the attributes. From this, any explanation method can be used in the attribute space, though this work has chosen to use SHAP~\cite{shap}.}
    \label{fig:attribute_expl}
\end{figure}

Outside of pixel importances, we are also interested in measuring the impact of a given attribute $\mathbf{a}_i$ on a classifier $f$. This problem has previously been briefly investigated by~\cite{dash2022evaluating}, the authors of the ImageCFGen causal generative model. However, the method from Dash et al. is limited to only binary classifiers and binary attributes $\mathbf{a}_i$. To make the notion of explanation formal, we again consider an observation $\mathbf{x,a}$ of an image and its attributes. We desire to have attribute importances for the values of $\mathbf{a}$, independent of the specific image $\mathbf{x}$ having these attributes (as there can be several images with the same attributes). Recall a causal generative model has a generator/decoder $G$, which can generate an image with the given attributes when a latent vector $\mathbf{z}\sim p(\mathbf{z})$ is supplied. As such, we define an approximate attribute classifier as the expected classification over all generated images with the given attributes. This is approximated by the following Monte carlo over a sample $\mathbf{z}_1,\mathbf{z}_2,\ldots,\mathbf{z}_m$ of size $m$ drawn from $p(\mathbf{z})$:
\begin{equation}
    \hat{f}(\mathbf{a}) = \mathbb{E}_{p(\mathbf{z})}[f(G(\mathbf{z},\mathbf{a}))] \approx \frac{1}{m}\sum_{i=1}^{m}{f(G(\mathbf{z}_i,\mathbf{a}))}.
\end{equation}
This process is described in~\autoref{fig:attribute_expl}. In our experiments, we set $m=4$. From this, we can visualize the SHAP values $\text{SHAP}(\mathbf{a};\hat{f})$ to determine feature importances in the attribute space. More specifically, we take the mean of absolute SHAP values over each of the $K$ class score functions $f_j$ to determine the importances of the given attribute. Further, to get a global feature importance, we consider the median value of the local feature importances accross the entire Morpho-MNIST test set for a given class (see~\autoref{fig:attribute_shap}).

\subsection{Counterfactual Explanations with Causal Generative Models}
\label{sec:cf_explanation_methods}

We propose two optimization approaches for creating counterfactual explainers using a generative causal model, which we refer to as our gradient-based and model-agnostic counterfactual explainers. As before, we denote $\mathbf{x}'(\mathbf{a}')$ as a counterfactual of a given observation $\mathbf{x,a}$ with the intervention $do(\mathbf{a}=\mathbf{a}')$. When searching over the digit label $\ell$, or generally any categorical attribute in $\mathbf{a}'$, we consider instead a random variable $\hat{\ell}$ taking possible values of $\ell$ (the digits 0 through 9 in this case) by representing it as a probability distribution over its values represented by a vector $\mathbf{p}$ with $\mathbf{p}_k = p(\hat{\ell} = k)$. We also assume that the categorical variable is passed to the generative models via an embedding lookup function $e(\ell)$. When a specific $\mathbf{p}$ (i.e., a specific random variable $\hat{\ell}$) is considered, it can be passed to a causal generative model by considering the following linear combination of embeddings:
\begin{equation}
    \label{eq:expected_label}
    \mathbb{E}_{p(\hat{\ell})}[e(\hat{\ell})] = \sum_{k=1}^{K}{\mathbf{p}_ke(k)}
\end{equation}
which is the expected value of the embedding vector $e(\cdot)$ following a transformation of the chosen distribution $\mathbf{p}$ over the categorical variabl $\ell$. This allows us to search over arbitrary mixtures of the categorical variable.

Our first approach uses a loss function similar to that of Watcher et al.'s  counterfactual explainer~\citep{wachter2018counterfactual}. OmnixAI modified the equation from Watcher et al., and provides a public-facing implementation based on the following formula:
\begin{equation}
   \label{eq:cf_example_omnix}    \min_{\mathbf{x}'}\max_\lambda{\lambda\mathcal{H}\left(f_y(\mathbf{x}') - \max_{y'\neq y} f_{y'}(\mathbf{x}')\right) + ||\mathbf{x}' - \mathbf{x}||_1}
\end{equation}
where $\mathcal{H}$ is the hinge loss function, $y$ is the predicted class of $\mathbf{x}$, $\mathbf{x}'$ is the counterfactual, $\lambda$ controls the tradeoff between the two loss terms, and $f_j$ is the classification score function for class $j$. However, in our gradient-based explainer approach, we account for a target class $y_t$ for the desired counterfactual, and search in the attribute space $\mathbf{a}'$. We propose the following loss function:
\begin{equation}
    \label{eq:cf_example_opt}
    \min_{\mathbf{a}'}\max_\lambda{\lambda\left(\max_{y'\neq y_t} f_{y'}(\mathbf{x}'(\mathbf{a}')) - f_{y_t}(\mathbf{x}'(\mathbf{a}'))\right) + ||\mathbf{x}'(\mathbf{a}') - \mathbf{x}||_1}.
\end{equation}
The above loss function can be restricted to a subset of attributes to further ensure that counterfactuals are visually close to the original image. In our experiments, we search only over the digit labels $\ell$, such that our counterfactual explanations lie somewhere between the space of images with the label $\ell=y$ and those with $\ell=y_t$. Further, we set $\lambda=10$ in our experiments rather than performing an explicit search after observing empirically that changing its value did not significantly impact performance. When searing over distributions $\mathbf{p}$ over the label $\ell$, we modify the logits of $\mathbf{p}$ by gradient descent on Formula~(\ref{eq:cf_example_opt}), passing these logits through a softmax before computing Equation~(\ref{eq:expected_label}). 

The second approach for counterfactual explanation we propose in this work is a model-agnostic optimization, which is based on a linear search. For a given classifier $f$, denote $C(\mathbf{x})$ as the most likely class predicted by $f$, i.e.:
\begin{equation}
    C(\mathbf{x}) = \arg\max_j f_j(\mathbf{x}).
\end{equation}
We use a form of Equation~(\ref{eq:expected_label}) to interpolate between images with the label $y=C(\mathbf{x})$ and the target label $y_t$, where $\mathbf{p}_{y_t} = \alpha, \mathbf{p}_y = 1-\alpha$, and $\mathbf{p}_k = 0$ for all other classes. We define an interpolated embedding $e_\alpha$ for $\alpha\in[0,1]$ as a linear combination of the embeddings for $y$ and $y_t$ as:
\begin{equation}
    e_\alpha = \alpha e(y_t) + (1-\alpha)e(y).
\end{equation}
Such that $e_\alpha$ equals $e(y)$ when $\alpha=0$ and $e(y_t)$ when $\alpha=1$. Such an approach is similar to well-known methods of interpolation in the latent space of generative models~\citep{liu2018data}, but occurs instead in the embedding space. Our goal is to find the smallest $\alpha$ such that, when we replace the embedding for $e(y)$ as the input to the counterfactual generator with $e_\alpha$ to generate a counterfactual $\mathbf{x}'$, we get a classification of $C(\mathbf{x}') = y_t$. This value of $\alpha$ is found in our experiments by searching over a grid of size 100. Because the proposed method does not use gradients (or even classification scores), it can be used on any image classifier, regardless of whether it is a neural network. However, in our experiments, we use a CNN classifier to allow for comparison with gradient-based methods.

\section{Experiments}

\begin{figure}[!t]
    \centering
    \includegraphics[width=0.7\textwidth]{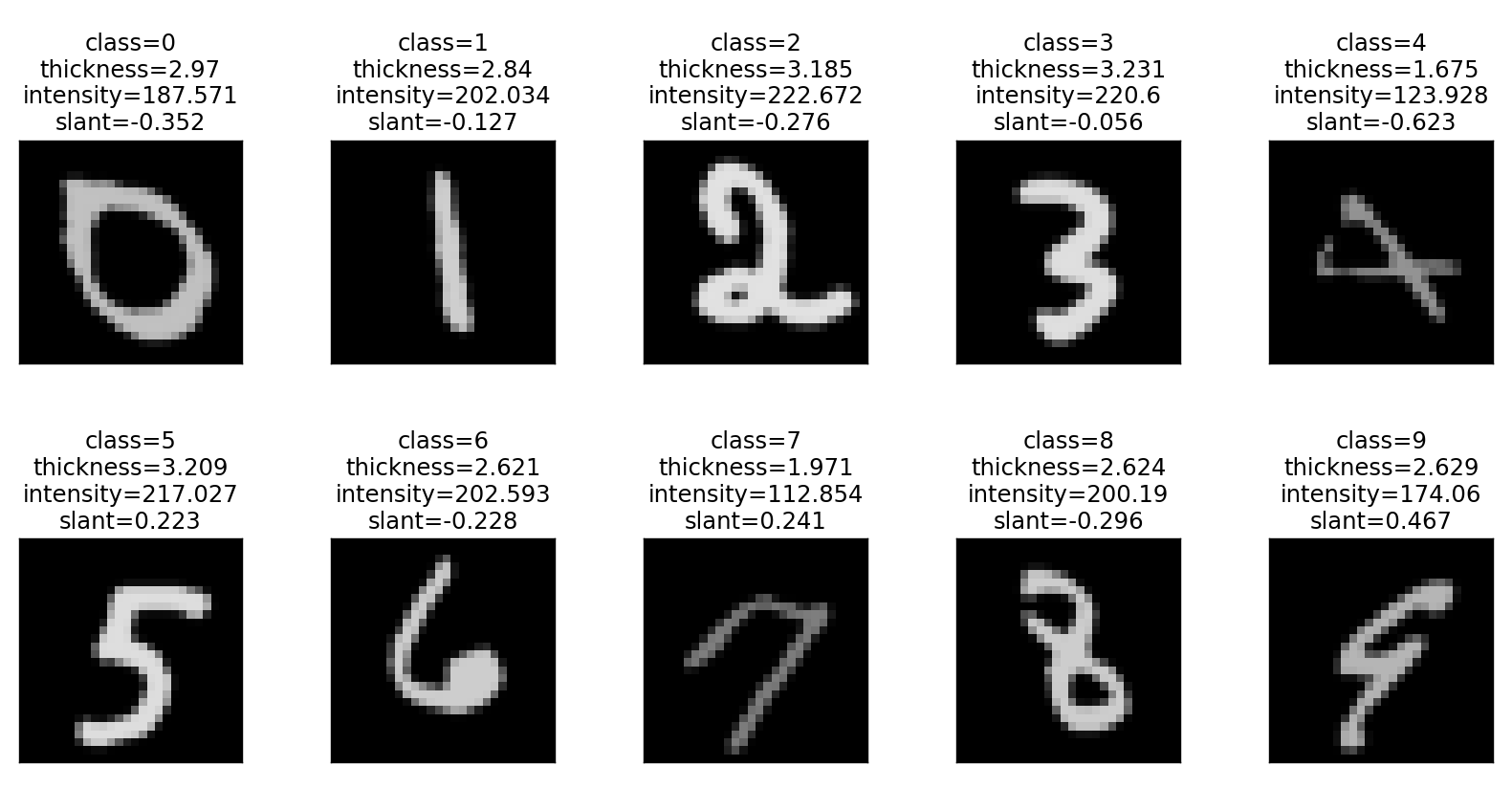}
    \caption{Instances from the Morpho-MNIST training set used in this work. Each digit has a varying class, thickness, intensity, and slant sampled from the ground-truth SCM.}
    \label{fig:morphomnist_data}
\end{figure}

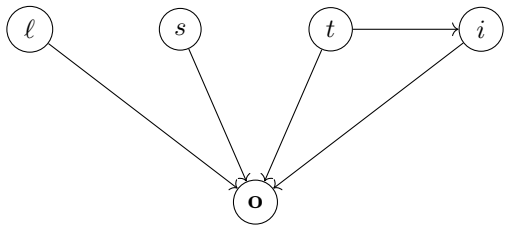
\begin{figure}[!t]
    \centering
    \begin{tikzpicture}[node distance=20mm, main/.style = {draw, circle}] 
        \node[main] (1) {$t$}; 
        \node[main] (2) [right of=1] {$i$}; 
        \node[main] (3) [left of=1] {$s$}; 
        \node[main] (4) [left of=3] {$\ell$}; 
        \node[main] (5) at ($(3)!0.5!(1)$) [below=2cm] {$\mathbf{o}$}; 
        \draw[->] (1) -- (2);
        \draw[->] (1) -- (5);
        \draw[->] (2) -- (5);
        \draw[->] (3) -- (5);
        \draw[->] (4) -- (5);
    \end{tikzpicture} 
    \caption{Causal graph for Morpho-MNIST, identical to the one used by~\cite{dash2022evaluating}.}
    \label{fig:morpho_mnist_graph}
\end{figure}

The Morpho-MNIST dataset~\citep{morpho-mnist} adds additional causal aspects to the well-known MNIST handwritten digits dataset~\citep{deng2012mnist}. Specifically, attributes of line thickness, image intensity, and slant are added to the original handwritten digits of MNIST. Because these attributes are added to the digits of MNIST using image processing techniques, scientists have complete control over the data generation process of Morpho-MNIST data and hence have full knowledge of the SCM used to generate Morpho-MNIST data.~\cite{pawlowski2020deep} were the first to create learned SCMs from the data generated using Morpho-MNIST techniques, and propose a causal graph involving the following attributes: 1. Thickness, denoted here as $t$. 2. Intensity, denoted here as $i$. 3. Digit image output, denoted here as $\mathbf{o}$.
We use in this work the variant of the Morpho-MNIST dataset used by~\cite{dash2022evaluating}, who added a slant parameter $s$ to the causal graph. The generative models are also conditioned on the digit label $\ell$. Instances from the Morpho-MNIST dataset are displayed in~\autoref{fig:morphomnist_data}, and the causal relationships between the variables of the dataset is shown in~\autoref{fig:morpho_mnist_graph}.

In all of our experiments, when training the VAE and BiGAN models of DeepSCM and ImageCFGen, the encoder and decoder for both the BiGAN and VAE consist of 5 convolutional layers (transposed convolutions in the decoder). The classifier being explained consists of 4 convolutional layers followed by a single fully-connected classification layer.

Throughout this section, we study the explainability of classifiers trained on the Morpho-MNIST dataset using the three methods which are described in~\autoref{sec:methods}. Moreover, we evaluate the quality of our counterfactual generation method from~\autoref{sec:cf_explanation_methods} using quantitative measurements and provide comparative results with an open-source framework for counterfactual generation.

\begin{figure}[!t]
    \centering
    \includegraphics[width=\textwidth]{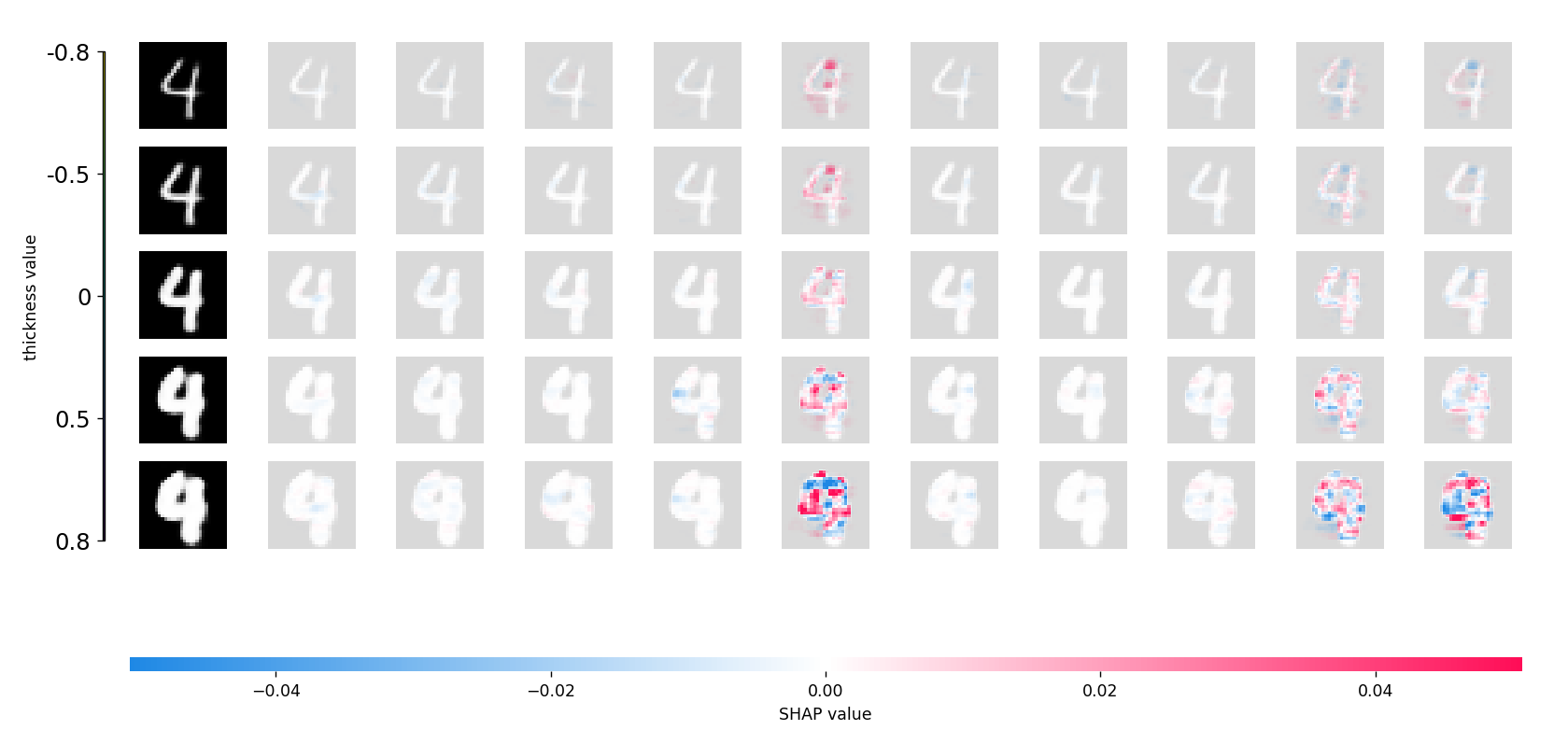}
    \caption{Evolution of SHAP pixel values as the thickness attribute of Morpho-MNIST varies on a digit `4'. Positive attributions (red) grow for the score for class 4 with the thickness, and positive regions for class 4 often correspond to negative regions for class 9 (and vice-versa). The displayed counterfactuals were computed using a VAE.}
    \label{fig:thickness_image_shap_vae}
\end{figure}

\begin{figure}[!ht]
    \centering
    \includegraphics[width=\textwidth]{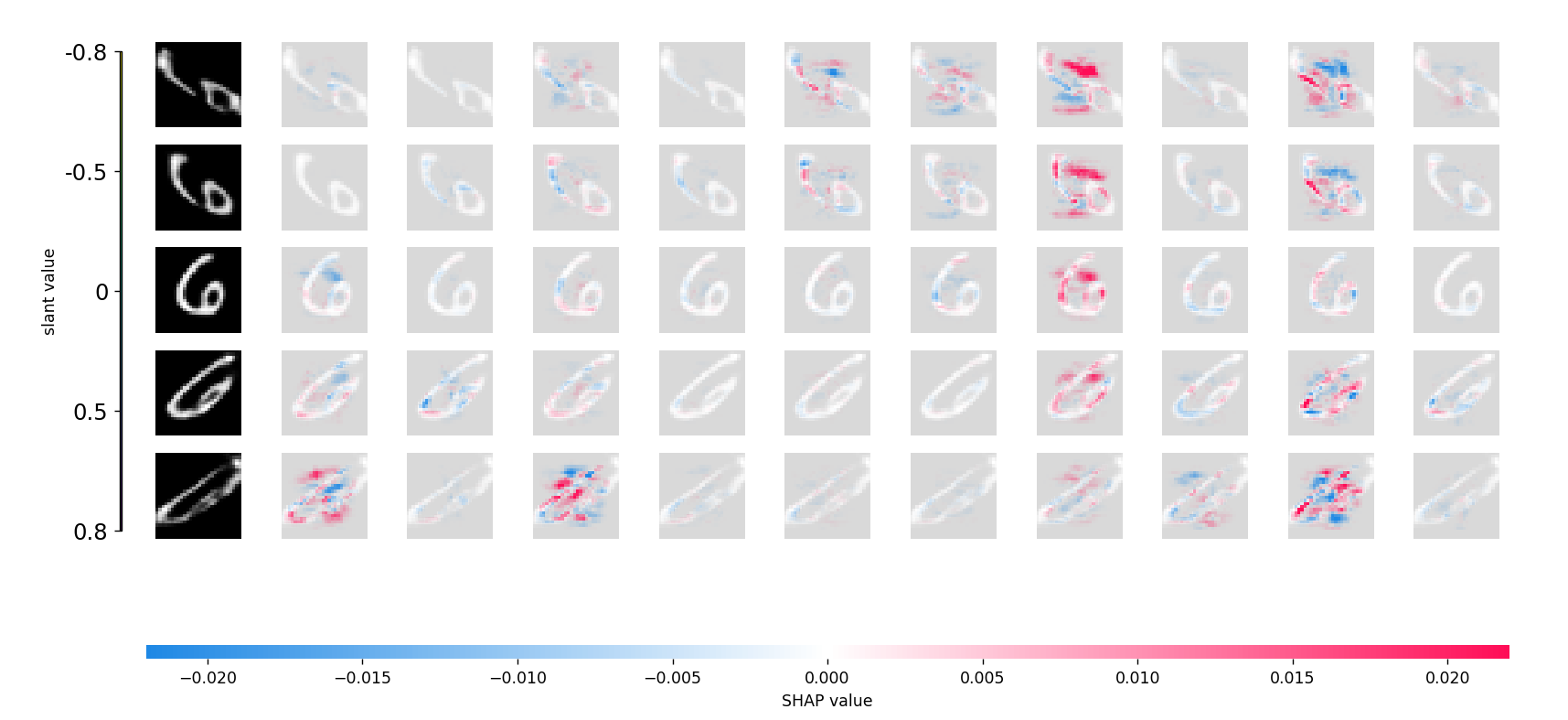}
    \caption{Evolution of SHAP pixel values as the slant attribute of Morpho-MNIST varies on a digit `6'.}
    \label{fig:slant_image_shap_vae}
\end{figure}

\subsection{Pixel Explanations}
We present the evolution of pixel importances for all three attributes measured by the SHAP explainer approach described in section~\autoref{sec:shap_contrastive_methods} using both VAE and BiGAN generative models. For both the SHAP and contrastive explainers, we normalize attributes to $[-1,1]$ in our experiments and set $S_i = \{-0.8, 0.5, 0, 0.5, 0.8\}$ for all attributes. We plot the image (classifier input) in the first column of our figures, with each subsequent column displaying the SHAP values for a specific digit class. We can observe in~\autoref{fig:thickness_image_shap_vae} that for lower levels of thickness values, the positive importance regions determined by SHAP are stronger and clearer. Further, comparing columns for class 4 and 9 in~\autoref{fig:thickness_image_shap_vae}, we can see that increasing thickness causes confusion between the two classes (positive and negative regions swap). We can also observe in~\autoref{fig:slant_image_shap_vae} that high magnitudes of slant make digits more difficult to classify. Specifically, on the digit 6 displayed, high clockwise slant leads to confusion with 0, 2, and 8, while high counter-clockwise slant leads to confusion with 4, 5, and 8.

\begin{figure}[!b]
    \centering
    \includegraphics[width=0.35\textwidth]{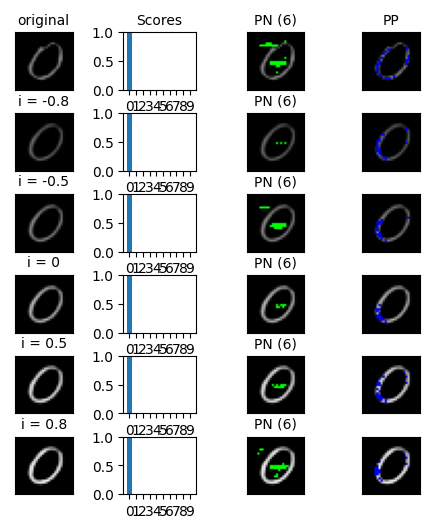}
    \caption{Contrastive explanations (PN and PP) from OmnixAI computed on counterfactuals with varying levels of intensity. Lower levels of intensity cause larger areas of PP to be required for the classification.}
    \label{fig:intensity_pp_pn}
\end{figure}

\begin{figure}[!ht]
    \centering
    \includegraphics[width=0.35\textwidth]{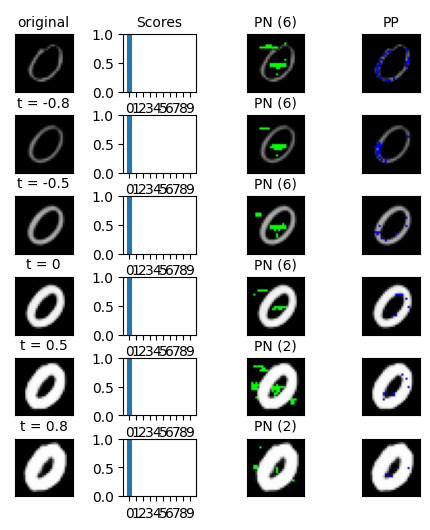}
    \caption{Contrastive explanations (PN and PP) from OmnixAI computed on counterfactuals with varying levels of thickness. Lower levels of thickness cause larger areas of PP to be required for the classification.}
    \label{fig:thickness_pp_pn}
\end{figure}

\begin{figure}[!htbp]
    \centering
    \includegraphics[width=0.35\textwidth]{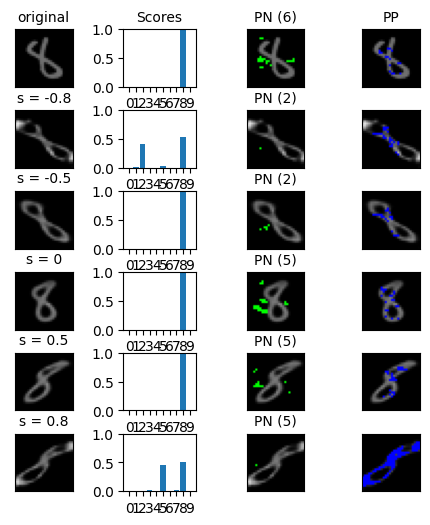}
    \caption{Very high blue PP area means that all features are necessary to avoid a misclassification on the extreme slant value. This suggests higher values of slant makes classification harder (more confusion).}
    \label{fig:slant_pp_pn}
\end{figure}

Contrastive explanations evolving with the values of intensity, thickness, and slant are shown in~\autoref{fig:intensity_pp_pn},~\autoref{fig:thickness_pp_pn}, and~\autoref{fig:slant_pp_pn}, respectively. We can see that with intensity (\autoref{fig:intensity_pp_pn}), no change in the attribute causes a change in the class scores from the classifier. However, smaller values of intensity are associated with larger PP regions, i.e., the high intensity images require less PP features to produce the same correct classification. The same phenomenon is seen with an increase in thickness (\autoref{fig:thickness_pp_pn}), however, because increases in thickness causes corresponding increases in intensity due to the causal structure of Morpho-MNIST (see~\autoref{fig:morpho_mnist_graph}), this observation is likely attributed to intensity rather than thickness. With slant (\autoref{fig:slant_pp_pn}), the highest value of slant causes all non-zero pixels to be part of the PP regions, suggesting again that slant can cause difficulty in classification as no smaller sub-region of the digit's features lead to a correct classification. This is also reflected in the classifiers score distribution, where the 8 with extreme positive slant is nearly incorrect classified as a 5.

\subsection{Attribute Explanation}

\begin{figure}[!htbp]
    \centering
    \includegraphics[width=.8\textwidth]{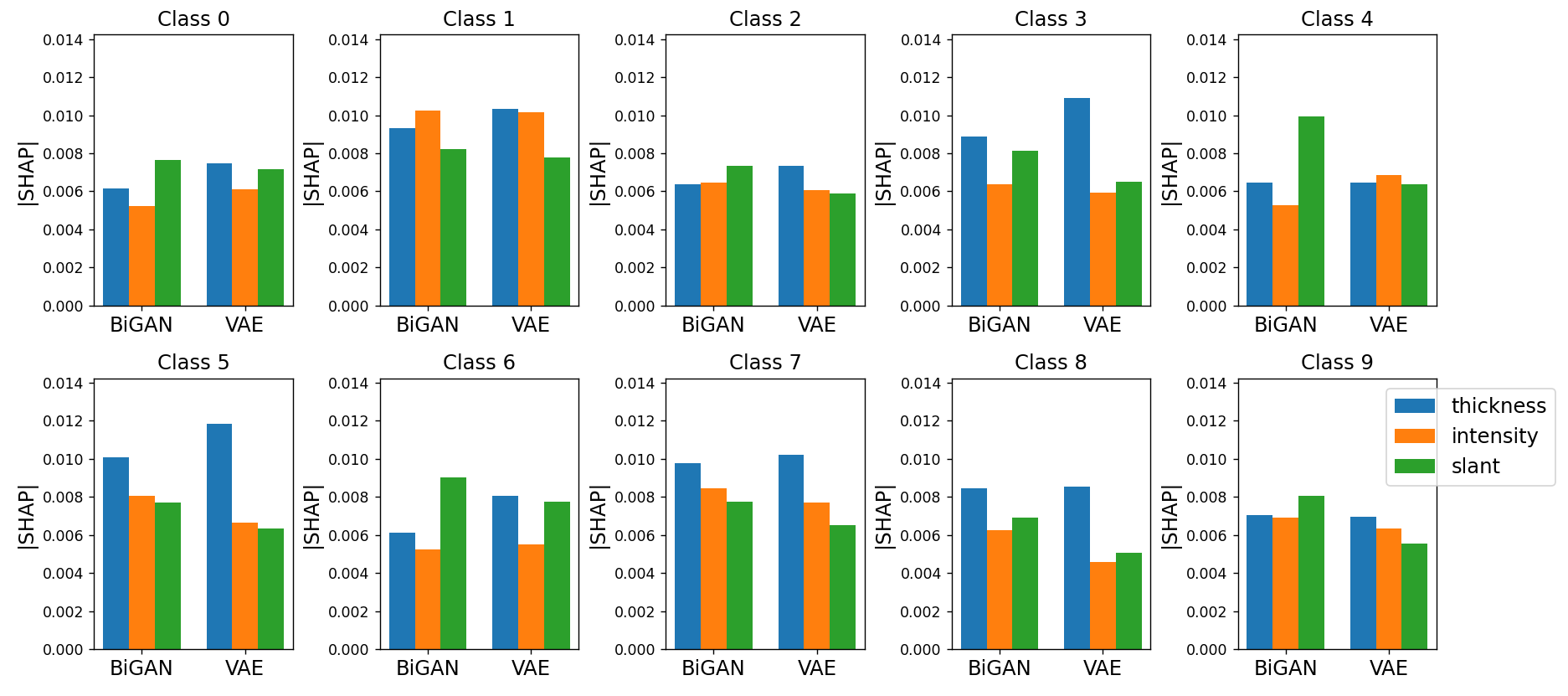}
    \caption{Attribute feature importances computed using the monte-carlo attribute classifier method (see~\autoref{fig:attribute_expl}), aggregated over all images from the classes the Morpho-MNIST test set as described in~\autoref{sec:shap_contrastive_methods}. Recall that thickness causes intensity (see~\autoref{fig:morpho_mnist_graph}).}
    \label{fig:attribute_shap}
\end{figure}

We also obtained the aggregate influence of each of the attributes on a classifier, which are depicted in~\autoref{fig:attribute_shap} for each of the ten classes of Morpho-MNIST. Our results suggest that while slant and thickness have the highest impact in changing digit classification, intensity plays a less evident role. Even though VAE and BiGAN models don't follow the same levels of attribute importances, they provide a similar order for ranking many of the most important attributes.

\subsection{Counterfactual Explanations}
To evaluate the interpretability of the counterfactual generation method proposed in~\autoref{sec:cf_explanation_methods}, and to compare the generated counterfactuals with those produced by OmnixAI explainers, we use the IM1 and IM2 scores from~\cite{van2021interpretable_im1} as well as the oracle score from~\cite{hvilshoj2021quantitative}. For an image $\mathbf{x}$ from class $p$ and corresponding counterfactual $\mathbf{x}'$ supposedly from the target class $q$, IM1 measures whether $\mathbf{x}'$ is closer to class $p$ or $q$ using autoencoders $\text{AE}_p$ and $\text{AE}_q$ trained on data coming from only the respective classes:
\begin{equation}
    \text{IM1} = \frac{||\mathbf{x}'-\text{AE}_q(\mathbf{x}')||_2^2}{||\mathbf{x}'-\text{AE}_p(\mathbf{x}')||_2^2 + \varepsilon}.
\end{equation}
If IM1 falls below 1, the counterfactual is better reconstructed by a model trained on data from the target class, suggesting it can be easily interpreted as an instance from the target class. IM2, also proposed by Van Looveren and Klaise, aims to measure how well the counterfactual follows the overall distribution of data by utilizing an autoencoder $\text{AE}$ trained on the entire training set:
\begin{equation}
    \text{IM2} = \frac{||\text{AE}_q(\mathbf{x}') - \text{AE}(\mathbf{x}')||_2^2}{||\mathbf{x}'||_1+\varepsilon}.
\end{equation}
Similarly to IM1, lower values of IM2 are considered better by the authors who proposed the metrics. Both the encoder and decoder of our autoencoders consist of two convolutional layers and one fully-connected layer.

\begin{figure}[!ht]
    \centering
    \includegraphics[width=.8\textwidth]{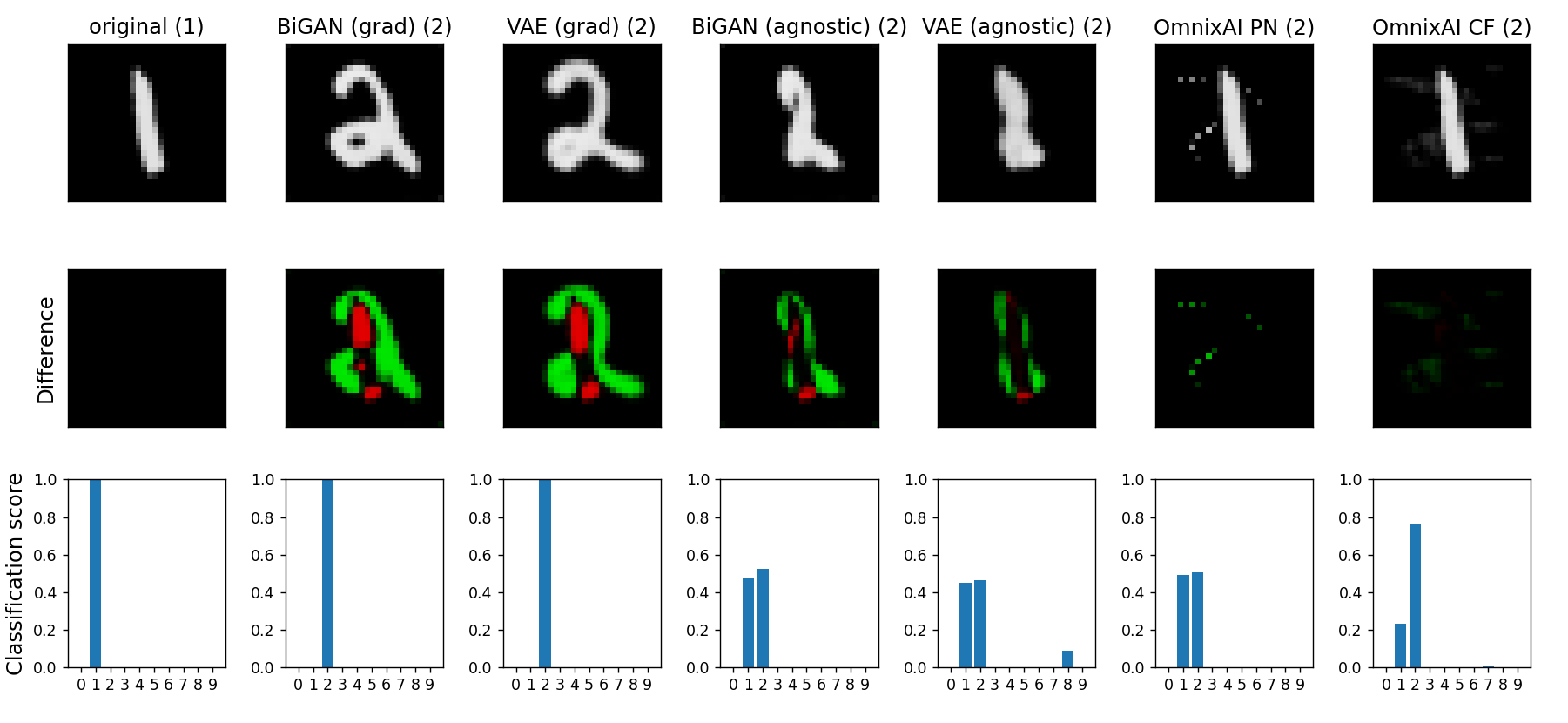}
    \caption{Comparison of visual explanation methods proposed in this work with the counterfactual and contrastive explainers from OmnixAI. The class of each explanation is shown in the title of each subfigure with the name of the explanation method, and the class score distribution from the classifier is shown underneath each image. The label ``OmnixAI PN" refers to the contrastive explainer. The required pixels to remove for converting an image of digit 1 to digit 2 are shown in red, and added pixels are shown in green. The third row indicates the accuracy of classification of the counterfactual.}
    \label{fig:cf_compare_1}
\end{figure}

\begin{figure}[!t]
    \centering
    \includegraphics[width=.9\textwidth]{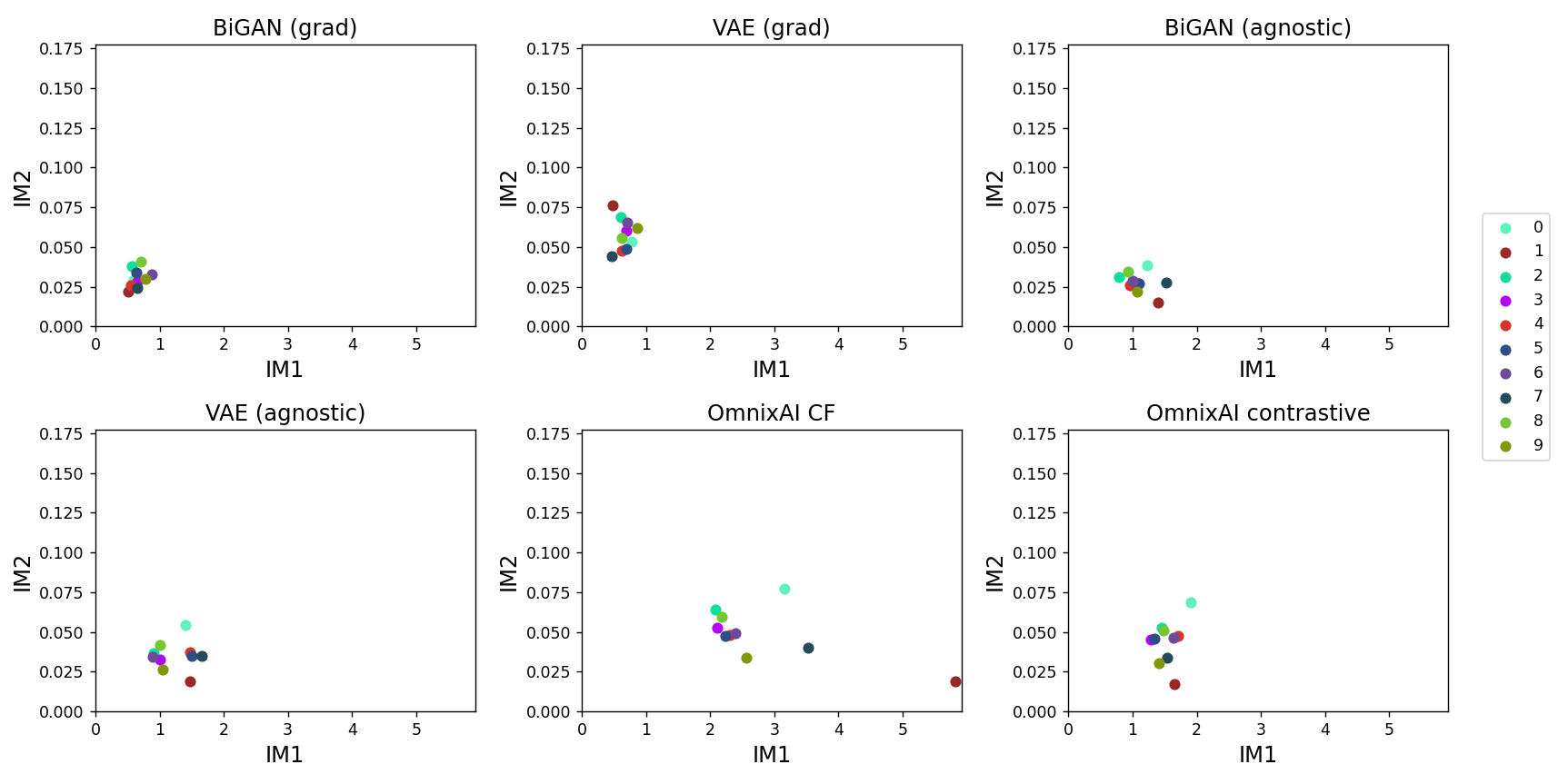}
    \caption{Scatter plots of mean IM1 and IM2 for each visual explanation method, plotted by Morpho-MNIST class (lower is better). The mean value for each class is shown in a different color.}
    \label{fig:im1_im2}
\end{figure}

\autoref{fig:im1_im2} compares the average distribution of IM1 vs. IM2 in a 2D scatter plot for all the ten classes from the test set of Morpho-MNIST. Values which are closer to the bottom-left corner indicate higher interpretability. As evident from the results, our approach achieves lower values of IM1 and IM2 in almost all cases. In addition, we can observe better consistency in terms of the variation of performance on different classes. It is also observable that both BiGAN and VAE models lead to similar results. However, as expected, our gradient-based method achieves better IM1 scores due to its tendency to produce embeddings close to the target class embedding $e(y_t)$ when compared with our model-agnostic approach. We note that although we report both IM1 and IM2 in~\autoref{fig:im1_im2}, IM2 has been previously been met with criticisms by other authors, with~\cite{schut2021_im2criticism} finding that IM2 fails to differentiate in-distribution from out-of-distribution images. Because of this, we report confidence intervals only for IM1 in~\autoref{tab:im1}. These results indicate that the IM1 values of our approaches are significantly better than those of OmnixAI. Overall, the results suggest that the best performance on these metrics is achieved by the gradient-based BiGAN method.

We also provide a visual example to compare the quality of counterfactually generated images of our methods with that of OmnixAI in~\autoref{fig:cf_compare_1}. Our approaches produce more visually interpretable images which are accurately classified as the target class.

\begin{table}[!t]
    \centering
    \begin{tabular}{|c|c|}
        \hline
        Method & IM1 95\% CI \\\hline
        BiGAN (grad) & \textbf{0.6489} $\pm$ 0.0073 \\\hline
        VAE (grad) & 0.6497 $\pm$ 0.0081 \\\hline
        BiGAN (agnostic) & 1.1124 $\pm$ 0.0104 \\\hline
        VAE (agnostic) & 1.2399 $\pm$ 0.0159 \\\hline
        OmnixAI CF & 2.8861 $\pm$ 0.0341 \\\hline
        OmnixAI contrastive & 1.5416 $\pm$ 0.0102 \\\hline
    \end{tabular}
    \caption{Recorded means and 95\% confidence intervals for IM1 scores for the visual explanation methods considered in this work, computed over the Morpho-MNIST test set. The lowest means are shown in \textbf{bold}.}
    \label{tab:im1}
\end{table}

\begin{figure}[!b]
    \centering
    \includegraphics[width=0.9\textwidth]{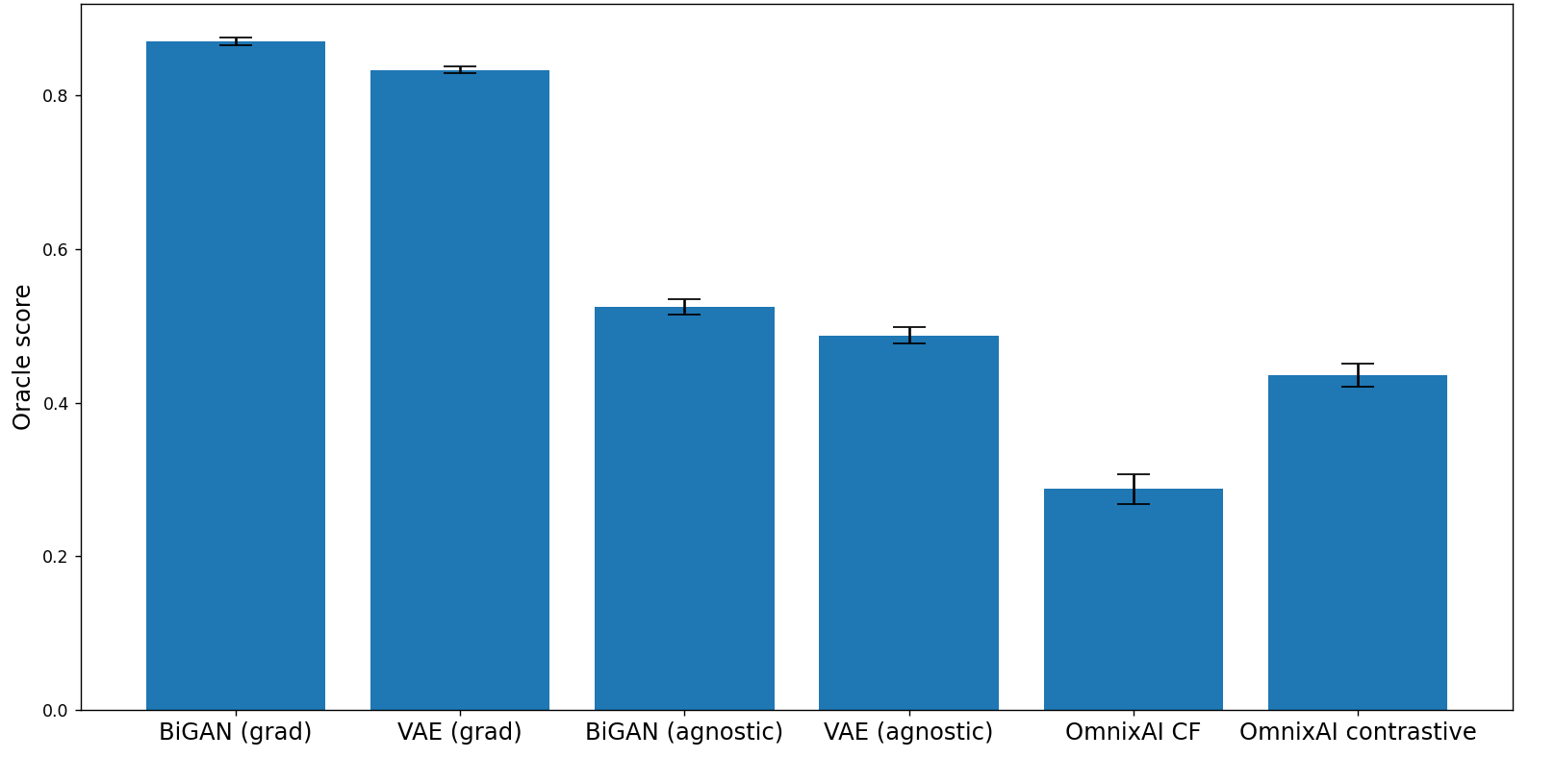}
    \caption{Oracle scores for each of the explanation methods considered in this work, computed across the Morpho-MNIST test set. 95\% confidence intervals for 10 runs with different oracle initializations are shown. All scores were significantly different at the 95\% confidence level, though the agnostic and contrastive explainers show similar performance on this metric.}
    \label{fig:oracle_score}
\end{figure}

The metrics proposed by Hvilshøj et al. are based on the presence of an \textit{oracle} $o$, which is an additional classifier trained with a different initial random state than the one used to train the explained classifier $f$.  Letting $\mathcal{X}'$ denote a set of counterfactuals explanations from a given method, the oracle score is defined as:
\begin{equation}
    \text{Oracle} = \frac{1}{|\mathcal{X}'|}\sum_{\mathbf{x}'\in\mathcal{X}'}{\mathbb{1}_{f(\mathbf{x}')=o(\mathbf{x}')}}
\end{equation}
where $\mathbb{1}_{f(\mathbf{x}')=o(\mathbf{x}')}$ is an indicator function with a value of 1 if $f$ and $o$ agree on the classification of counterfactual and 0 otherwise. The idea behind this metric is that if the classifier and oracle do not agree on many counterfactuals, then the changes made to the original instances are very specific to the weights of $f$ (i.e., they are adversarial and not interpretable). ~\autoref{fig:oracle_score} presents the mean oracle scores over 10 independent runs for all the explainers, using different oracles on each run. The results are consistent with our previous observation that our proposed explainers provide superior interpretability in comparison to OmnixAI. In particular, we note that our gradient-based explainer outperforms all the methods which are discussed in this paper.

\section{Discussion and Conclusion}
This paper presented methods by which causal generative models can be used to explain image classifiers through counterfactual inference. These explanations fell into three main categories. The first method used counterfactuals with varying levels of attributes to analyze the variation in pixel-based explanations, using both SHAP values and contrastive explainers. The second method used a modification to the standard method of extracting SHAP values to extract attribute importances. The third method proposed both gradient-based and model-agnostic counterfactual explanations of image classifiers by searching in the attribute space of a causal generative model.

Our method of analyzing pixel-based explanations for varying levels of a causal attribute (\autoref{sec:shap_contrastive_methods}) helps us identify which levels of an attribute can lead to near-misclassifications or confusion with other classes, as well as determine which levels of an attribute can lead to simpler classifications due to a need for less features to form a pertinent positive.

Our method of identifying attribute importances from~\autoref{sec:attribute_explanations} is able to rank the human-interpretable features of a causal dataset in terms of their importance to an image classifier. This method expands upon the counterfactual importance score from~\cite{dash2022evaluating}, in that it works for arbitrary classifiers and arbitrary attributes rather than only binary attributes and binary classifiers.

The methods presented in this work for the generation of counterfactual explanations are demonstrably more interpretable than those from OmnixAI. We showed this using quantitative metrics, as well as by displaying examples of visual explanations. Further, these methods can be deployed using existing causal generative models, eliminating the need for additional training of specialized explanation models as is required by current generative model-based explanation methods.

One possible application of our proposed method is applying our proposed attribute feature explainers to the legibility of one's handwriting. The explainability features derived from our methods could in theory be embedded in a recommendation system to give feedback to the users in order to improve the readability level of handwriting. We also suggest investigating new loss functions to use with our gradient-based counterfactual explainers in future work, possibly incorporating more advanced regularization techniques or incorporating a GAN's discriminator in the loss function to further ensure the generation of realistic counterfactual examples. Additionally, for our model-agnostic explainer, more advanced interpolation techniques could be investigated rather than the simple linear interpolation used in this work.

\section*{Acknowledgements}
\noindent This work was supported by the Natural Sciences and Engineering Research Council of Canada [grant number 550722].

\bibliographystyle{unsrtnat}
\bibliography{references}

\begin{thebibliography}{26}
\providecommand{\natexlab}[1]{#1}
\providecommand{\url}[1]{\texttt{#1}}
\expandafter\ifx\csname urlstyle\endcsname\relax
  \providecommand{\doi}[1]{doi: #1}\else
  \providecommand{\doi}{doi: \begingroup \urlstyle{rm}\Url}\fi

\bibitem[Rezende and Mohamed(2015)]{rezende2015variational}
Danilo Rezende and Shakir Mohamed.
\newblock Variational inference with normalizing flows.
\newblock In \emph{International conference on machine learning}, pages 1530--1538. PMLR, 2015.

\bibitem[Goodfellow et~al.(2020)Goodfellow, Pouget-Abadie, Mirza, Xu, Warde-Farley, Ozair, Courville, and Bengio]{goodfellow2020generative}
Ian Goodfellow, Jean Pouget-Abadie, Mehdi Mirza, Bing Xu, David Warde-Farley, Sherjil Ozair, Aaron Courville, and Yoshua Bengio.
\newblock Generative adversarial networks.
\newblock \emph{Communications of the ACM}, 63\penalty0 (11):\penalty0 139--144, 2020.

\bibitem[Pawlowski et~al.(2020)Pawlowski, Coelho~de Castro, and Glocker]{pawlowski2020deep}
Nick Pawlowski, Daniel Coelho~de Castro, and Ben Glocker.
\newblock Deep structural causal models for tractable counterfactual inference.
\newblock \emph{Advances in Neural Information Processing Systems}, 33:\penalty0 857--869, 2020.

\bibitem[Dash et~al.(2022)Dash, Balasubramanian, and Sharma]{dash2022evaluating}
Saloni Dash, Vineeth~N Balasubramanian, and Amit Sharma.
\newblock Evaluating and mitigating bias in image classifiers: A causal perspective using counterfactuals.
\newblock In \emph{Proceedings of the IEEE/CVF Winter Conference on Applications of Computer Vision}, pages 915--924, 2022.

\bibitem[Schölkopf and von Kügelgen(2022)]{scholkopf-statistical-to-causal}
Bernhard Schölkopf and Julius von Kügelgen.
\newblock From statistical to causal learning, 2022.
\newblock URL \url{https://arxiv.org/abs/2204.00607}.

\bibitem[Dwivedi et~al.(2023)Dwivedi, Dave, Naik, Singhal, Omer, Patel, Qian, Wen, Shah, Morgan, et~al.]{dwivedi2023explainable}
Rudresh Dwivedi, Devam Dave, Het Naik, Smiti Singhal, Rana Omer, Pankesh Patel, Bin Qian, Zhenyu Wen, Tejal Shah, Graham Morgan, et~al.
\newblock Explainable ai (xai): Core ideas, techniques, and solutions.
\newblock \emph{ACM Computing Surveys}, 55\penalty0 (9):\penalty0 1--33, 2023.

\bibitem[Van~Looveren et~al.(2021)Van~Looveren, Klaise, Vacanti, and Cobb]{van2021conditionalcf}
Arnaud Van~Looveren, Janis Klaise, Giovanni Vacanti, and Oliver Cobb.
\newblock Conditional generative models for counterfactual explanations.
\newblock \emph{arXiv preprint arXiv:2101.10123}, 2021.

\bibitem[Yang et~al.(2021)Yang, Alva, Chen, and Hu]{yang2021cfmodel}
Fan Yang, Sahan~Suresh Alva, Jiahao Chen, and Xia Hu.
\newblock Model-based counterfactual synthesizer for interpretation.
\newblock In \emph{Proceedings of the 27th ACM SIGKDD conference on knowledge discovery \& data mining}, pages 1964--1974, 2021.

\bibitem[Lundberg and Lee(2017)]{shap}
Scott~M Lundberg and Su-In Lee.
\newblock A unified approach to interpreting model predictions.
\newblock In I.~Guyon, U.~V. Luxburg, S.~Bengio, H.~Wallach, R.~Fergus, S.~Vishwanathan, and R.~Garnett, editors, \emph{Advances in Neural Information Processing Systems 30}, pages 4765--4774. Curran Associates, Inc., 2017.
\newblock URL \url{http://papers.nips.cc/paper/7062-a-unified-approach-to-interpreting-model-predictions.pdf}.

\bibitem[Dhurandhar et~al.(2018)Dhurandhar, Chen, Luss, Tu, Ting, Shanmugam, and Das]{dhurandhar2018contrastive}
Amit Dhurandhar, Pin-Yu Chen, Ronny Luss, Chun-Chen Tu, Paishun Ting, Karthikeyan Shanmugam, and Payel Das.
\newblock Explanations based on the missing: Towards contrastive explanations with pertinent negatives, 2018.

\bibitem[Kobyzev et~al.(2020)Kobyzev, Prince, and Brubaker]{kobyzev2020normalizing}
Ivan Kobyzev, Simon~JD Prince, and Marcus~A Brubaker.
\newblock Normalizing flows: An introduction and review of current methods.
\newblock \emph{IEEE transactions on pattern analysis and machine intelligence}, 43\penalty0 (11):\penalty0 3964--3979, 2020.

\bibitem[de~Castro et~al.(2018)de~Castro, Tan, Kainz, Konukoglu, and Glocker]{morpho-mnist}
Daniel~Coelho de~Castro, Jeremy Tan, Bernhard Kainz, Ender Konukoglu, and Ben Glocker.
\newblock Morpho-mnist: Quantitative assessment and diagnostics for representation learning.
\newblock \emph{CoRR}, abs/1809.10780, 2018.
\newblock URL \url{http://arxiv.org/abs/1809.10780}.

\bibitem[Deng(2012)]{deng2012mnist}
Li~Deng.
\newblock The mnist database of handwritten digit images for machine learning research.
\newblock \emph{IEEE Signal Processing Magazine}, 29\penalty0 (6):\penalty0 141--142, 2012.

\bibitem[Donahue et~al.(2016)Donahue, Kr{\"a}henb{\"u}hl, and Darrell]{bigan}
Jeff Donahue, Philipp Kr{\"a}henb{\"u}hl, and Trevor Darrell.
\newblock Adversarial feature learning.
\newblock \emph{arXiv preprint arXiv:1605.09782}, 2016.

\bibitem[Dumoulin et~al.(2016)Dumoulin, Belghazi, Poole, Mastropietro, Lamb, Arjovsky, and Courville]{ali}
Vincent Dumoulin, Ishmael Belghazi, Ben Poole, Olivier Mastropietro, Alex Lamb, Martin Arjovsky, and Aaron Courville.
\newblock Adversarially learned inference.
\newblock \emph{arXiv preprint arXiv:1606.00704}, 2016.

\bibitem[Feng et~al.(2021)Feng, Lin, Zhu, Zhao, Zhou, and Zha]{feng2021uncertainty}
Ruili Feng, Zhouchen Lin, Jiapeng Zhu, Deli Zhao, Jingren Zhou, and Zheng-Jun Zha.
\newblock Uncertainty principles of encoding gans.
\newblock In \emph{International Conference on Machine Learning}, pages 3240--3251. PMLR, 2021.

\bibitem[Parafita and Vitri{\`a}(2019)]{parafita2019explaining}
{\'A}lvaro Parafita and Jordi Vitri{\`a}.
\newblock Explaining visual models by causal attribution.
\newblock In \emph{2019 IEEE/CVF International Conference on Computer Vision Workshop (ICCVW)}, pages 4167--4175. IEEE, 2019.

\bibitem[Pearl(2009)]{pearl2009causality}
Judea Pearl.
\newblock \emph{Causality}.
\newblock Cambridge university press, 2009.

\bibitem[Hvilsh{\o}j et~al.(2021{\natexlab{a}})Hvilsh{\o}j, Iosifidis, and Assent]{hvilshoj2021ecinn}
Frederik Hvilsh{\o}j, Alexandros Iosifidis, and Ira Assent.
\newblock Ecinn: efficient counterfactuals from invertible neural networks.
\newblock \emph{arXiv preprint arXiv:2103.13701}, 2021{\natexlab{a}}.

\bibitem[Dinh et~al.(2014)Dinh, Krueger, and Bengio]{dinh2014nice}
Laurent Dinh, David Krueger, and Yoshua Bengio.
\newblock Nice: Non-linear independent components estimation.
\newblock \emph{arXiv preprint arXiv:1410.8516}, 2014.

\bibitem[Mahajan et~al.(2019)Mahajan, Tan, and Sharma]{mahajan2019preservingcf}
Divyat Mahajan, Chenhao Tan, and Amit Sharma.
\newblock Preserving causal constraints in counterfactual explanations for machine learning classifiers.
\newblock \emph{arXiv preprint arXiv:1912.03277}, 2019.

\bibitem[Wachter et~al.(2018)Wachter, Mittelstadt, and Russell]{wachter2018counterfactual}
Sandra Wachter, Brent Mittelstadt, and Chris Russell.
\newblock Counterfactual explanations without opening the black box: Automated decisions and the gdpr, 2018.

\bibitem[Liu et~al.(2018)Liu, Zou, Kong, Diao, Yan, Wang, Li, Jia, and You]{liu2018data}
Xiaofeng Liu, Yang Zou, Lingsheng Kong, Zhihui Diao, Junliang Yan, Jun Wang, Site Li, Ping Jia, and Jane You.
\newblock Data augmentation via latent space interpolation for image classification.
\newblock In \emph{2018 24th International Conference on Pattern Recognition (ICPR)}, pages 728--733. IEEE, 2018.

\bibitem[Van~Looveren and Klaise(2021)]{van2021interpretable_im1}
Arnaud Van~Looveren and Janis Klaise.
\newblock Interpretable counterfactual explanations guided by prototypes.
\newblock In \emph{Joint European Conference on Machine Learning and Knowledge Discovery in Databases}, pages 650--665. Springer, 2021.

\bibitem[Hvilsh{\o}j et~al.(2021{\natexlab{b}})Hvilsh{\o}j, Iosifidis, and Assent]{hvilshoj2021quantitative}
Frederik Hvilsh{\o}j, Alexandros Iosifidis, and Ira Assent.
\newblock On quantitative evaluations of counterfactuals.
\newblock \emph{arXiv preprint arXiv:2111.00177}, 2021{\natexlab{b}}.

\bibitem[Schut et~al.(2021)Schut, Key, Mc~Grath, Costabello, Sacaleanu, Gal, et~al.]{schut2021_im2criticism}
Lisa Schut, Oscar Key, Rory Mc~Grath, Luca Costabello, Bogdan Sacaleanu, Yarin Gal, et~al.
\newblock Generating interpretable counterfactual explanations by implicit minimisation of epistemic and aleatoric uncertainties.
\newblock In \emph{International Conference on Artificial Intelligence and Statistics}, pages 1756--1764. PMLR, 2021.

\end{thebibliography}

\end{document}